\documentclass[10pt,twocolumn,letterpaper]{article}

\usepackage[pagenumbers]{wacv} 
\newlength{\rewardfigheight}
\usepackage{graphicx}
\usepackage{booktabs}
\usepackage[table]{xcolor}
\usepackage{stfloats}
\usepackage{placeins}

\definecolor{best}{RGB}{255,180,90}     
\definecolor{second}{RGB}{255,220,170}  

\definecolor{wacvblue}{rgb}{0.21,0.49,0.74}
\usepackage[pagebackref,breaklinks,colorlinks,allcolors=wacvblue]{hyperref}

\def\wacvPaperID{1935} 
\def\confName{WACV}
\def\confYear{2027}

\title{Can Vision-Language Models Reason about AI Edits in Images?}

\author{
Darsha Udayanga$^{1}$ \and
Pin-Yu Chen$^{2}$ \and
Payel Das$^{2}$ \and
Qiang Ji$^{1}$\\
$^{1}$Rensselaer Polytechnic Institute \qquad
$^{2}$IBM Research\\
{\tt\small heratd@rpi.edu}
\qquad
{\tt\small pin-yu.chen@ibm.com}\\
{\tt\small daspa@us.ibm.com}
\qquad
{\tt\small jiq@rpi.edu}
}

\begin{document}
\maketitle
\begin{abstract}
Detection and localization of AI-tampered images are critical for trustworthy AI, yet modern generative models have made such manipulations increasingly difficult to identify. While traditional binary classifiers can detect image tampering, they lack interpretability and generalization. Vision-Language Models (VLMs) offer a promising alternative due to their strong visual understanding and reasoning capabilities; however, existing approaches typically rely on supervised finetuning with curated explanations rather than exploiting their inherent reasoning capabilities. In this work, we investigate whether VLMs can be trained to reason about AI-generated image edits using reinforcement learning (RL) rather than explicit reasoning supervision. Motivated by the success in Group Relative Policy Optimization (GRPO), an RL technique that incentivizes the model to reason by generating thinking traces prior to giving the final answer, we propose a GRPO-based training framework that utilizes simple accuracy and format rewards. Given an input image, the model produces a structured reasoning trace and predicts whether the image has been tampered with. A lightweight segmentation model is then guided by the reasoning output to generate pixel-level localization masks. Experiments across multiple image manipulation datasets demonstrate that our approach achieves competitive detection and localization performance compared to state-of-the-art image forgery detectors, despite requiring substantially weaker supervision. We introduce effective intersection over union (eff-IoU), a unified metric to jointly evaluate detection and localization. These results suggest that reinforcement learning provides an effective and scalable mechanism for teaching VLMs to reason about AI-generated content.
\end{abstract}

\begin{figure*}[t]
    \centering
    \includegraphics[width=\textwidth]{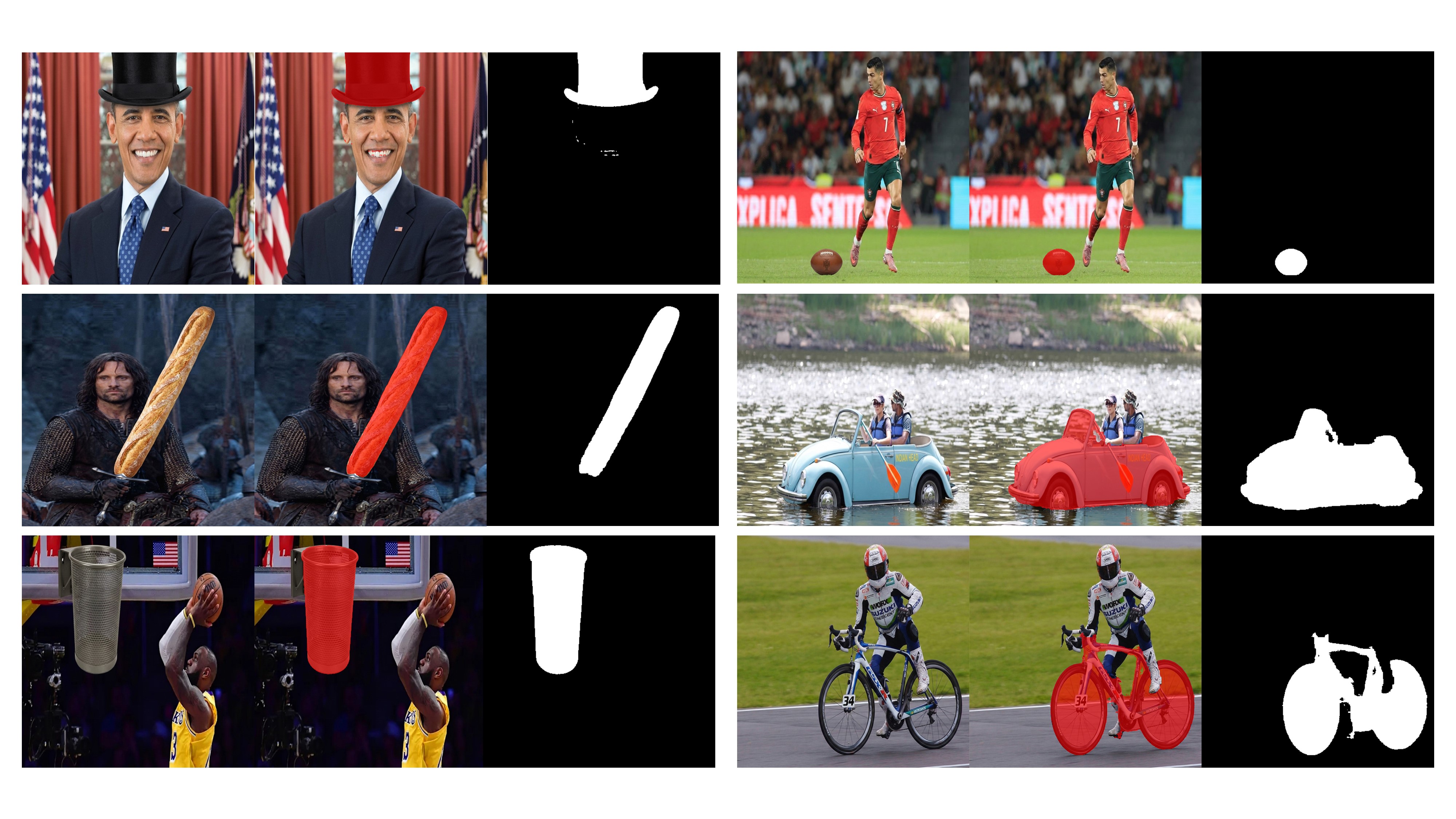}
    \caption{Examples of realistic image manipulations. 
    For each example, the columns show (left to right): the manipulated image, the manipulated image with the tampered region highlighted in red, and the corresponding predicted binary tamper mask. The presented results are generated by the method proposed in this paper.}
    \label{fig:fig_1}
\end{figure*}

\section{Introduction}
\label{sec:intro}

With the rapid development of Artificial Intelligence (AI), the debate has arisen over its societal impact. While large and widely accessible generative models have positively influenced many aspects of daily life, they have also enabled misuse in areas such as fraud and privacy infringement. Compared to a few years ago, machine-generated content has become increasingly difficult to distinguish from authentic content. Large Language Models (LLMs), such as ChatGPT\cite{gpt3} and Gemini\cite{gemini}, have demonstrated remarkable capabilities in text generation, while significant progress in image and video synthesis has been driven by advances in diffusion-based models\cite{diffusion_1,diffusion_2}.

The realistic nature of this content has opened new avenues for misuse and has made detection increasingly challenging for the average user. Examples include deepfake impersonation, document forgery, fake news generation, and financial scams. This has motivated growing research efforts in trustworthy AI, including AI-generated content (AIGC) detection, watermarking, and content authentication \cite{forgery_det_1,forgery_det_2}. At the same time, an adversarial dynamic has emerged, where generative models and detection methods continuously evolve to outperform one another, making trustworthy AI an ongoing challenge.

Existing detection methods range from metadata analysis \cite{metadata_1,metadata_2}, watermark detection\cite{watermark}, spectral analysis\cite{spectrum}, and lighting-based cues\cite{lighting} to deep-learning-based approaches such as multimodal large language models (MLLMs). MLLMs\cite{forgegpt} leverage both visual and textual information to reason about potential image edits, offering improved robustness and generalization compared to traditional methods, albeit with susceptibility to hallucinations\cite{hallu_1,hallu_2}. Recent work has focused on providing explanations alongside predictions, leading to the development of explainable Image Forgery Detection and Localization (e-IFDL) methods. However, most state-of-the-art approaches rely on strongly supervised training with curated reasoning annotations and often generalize poorly beyond their training distributions \cite{ImageManipulationDetection,HierarchicalFine-Grained}. Motivated by recent success in teaching an LLM to think to reason and the observed significant advances in areas such as mathematical problem solving and coding, we ask the following foundational Research Questions (RQs) with the objective to bridge the gap between VLM reasoning and tamper detection.

\begin{enumerate}
     \item \textbf{RQ1: Can we design a novel reward model that will enable reasoning (i.e., thinking before answering) to detect tampered images?}
     \\
    \item \textbf{RQ2: Can such a framework generalize across different VLM architectures and datasets?}
    \\
    \item \textbf{RQ3: Can weakly supervised reasoning-based models achieve performance comparable to state-of-the-art detectors trained with strong supervision?}
\end{enumerate}



Our approach to these questions is to develop a generalized reasoning model while leveraging the vision capabilities of a Vision Language Model (VLM). Our method was motivated by the recent utilization of Group Relative Policy Optimization (GRPO) for DeepSeek R1 \cite{guo_deepseek-r1_2025}. GRPO is a Reinforcement Learning (RL) technique where the model generates multiple generations for the same input and compares them against each other to guide the model to learn. The comparison is done via a set of reward functions that evaluate different criteria of the generation. In the proposed study, the VLM takes an image as input and generates a reasoning trace on whether the image is tampered with or not, and why. Then, a segmentation model extracts the edited segments following the reasoning trace. Our final model output is two-fold. The binary classification, where the model decides if the image is tampered or not, and the mask of the edited regions provided by the segmentation model, if there is any. In addition to that, the reasoning flow for making the decision is provided to the user as well. We have provided some results from real-world instances in Fig. \ref{fig:fig_1}.

To answer \textbf{Research Question 1}, we carefully designed simple but effective reward functions that are capable of guiding the VLM to reason and produce an answer based on that reasoning. To address \textbf{Research Question 2}, the framework was applied on few different models and different datasets. Unlike most studies, we experimented with and without authentic images to investigate whether the model can learn from both types of data, authentic and tampered, to develop the reasoning process. Finally, for the \textbf{Research Question 3}, the trained models were evaluated against SOTA studies using different metrics, such as classification accuracy and localization accuracy, which proved our technique performs competitively with SOTA models despite the weak supervision. Additionally, we introduce a new metric named effective-IoU as a unified metric for jointly evaluating classification and localization accuracies. 

Our proposed RL training methods show promising paths to enhance the capability in teaching VLMs to reason about AI edits in images. Following the trajectory of AI reasoning in LLMs enabled by GRPO in general domains, 
we expect that our findings will serve as the foundation of future studies to improve the capability in detecting and understanding hidden traces caused AI edits, through means such as scaling up training data volume or model size, among others.

\begin{figure*}[t]
    \centering
    \includegraphics[width=\textwidth]{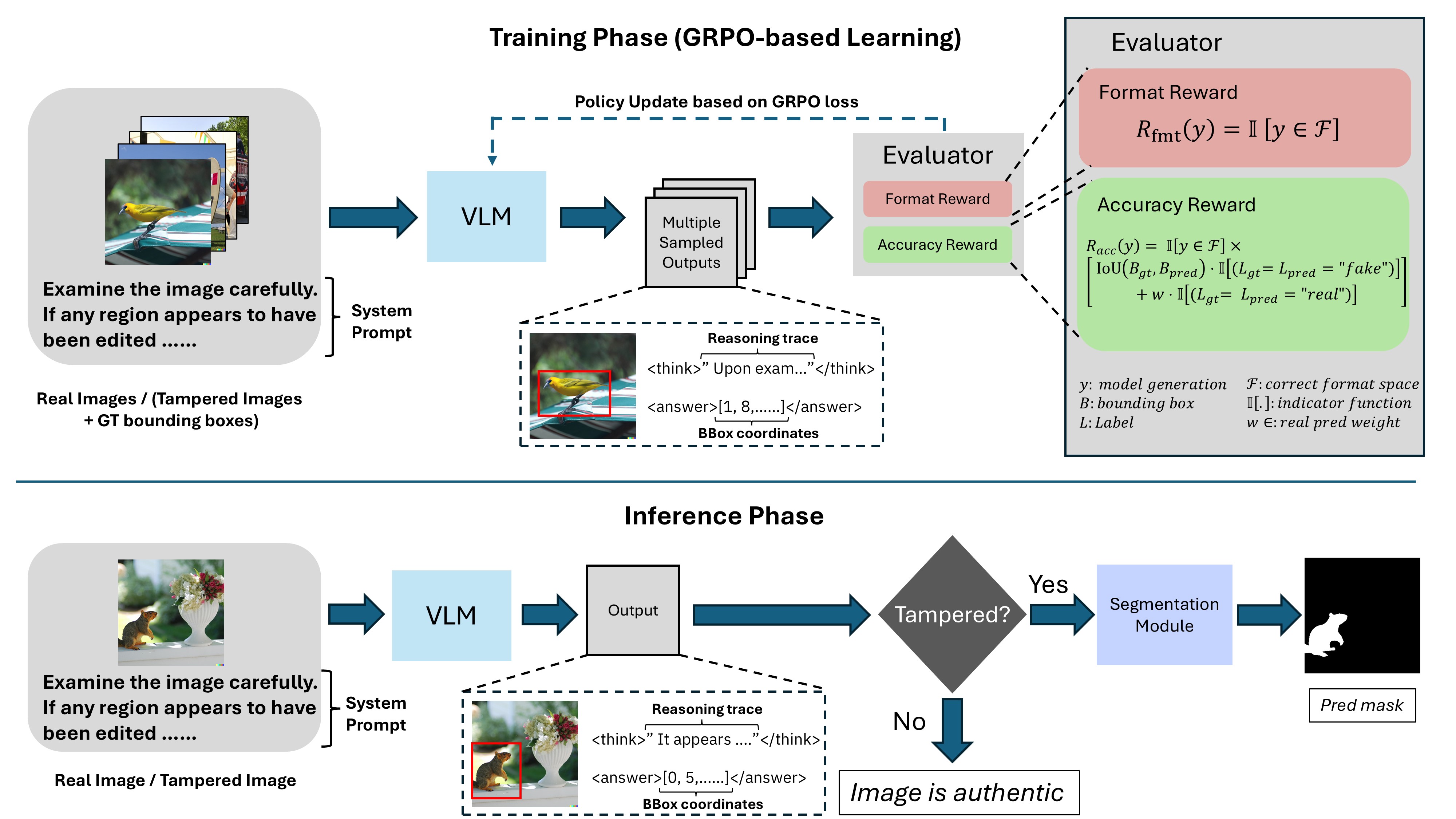}
    \caption{Overview of the proposed framework. 
    \textbf{Training phase (top):} The VLM receives an image and generates multiple reasoning-based outputs. 
    These generations are evaluated using format and accuracy rewards, and the model is updated using the GRPO loss. 
    \textbf{Inference phase (bottom):} The VLM predicts if an image is tampered and produces a reasoning trace with a bounding box. 
    If tampering is detected, a segmentation module refines the predicted region to produce the final tamper mask. Else, predicted as an authentic image.}
    \label{fig:overall}
\end{figure*}

\section{Background and Related Work}

\subsection{Group Relative Policy Optimization (GRPO)}
One drawback of traditional Supervised Finetuning (SFT) is that it requires large labeled datasets, where a popular alternative is Reinforcement Learning. An RL model learns by interacting with its environment, and GRPO is one RL technique. 
Essentially, it generates multiple responses from the same model for the same input and evaluates them against each other. The core of GRPO is this multi-output evaluation and it is carried out through carefully designed reward functions. The GRPO loss is calculated from these comparisons and used to optimize the model. The main advantage is that it removes the need for a separate critique model as in general RL.

The optimization objective for GRPO (GRPO loss) is given in equation \ref{eq:grpo}. This compares the relative advantage within the group per generation and weights the gradients. 
\begin{equation}
\begin{aligned}
J_{\mathrm{GRPO}}(\theta)
=&
\frac{1}{G}
\sum_{i=1}^{G}
\frac{1}{|o_i|}
\sum_{t=1}^{|o_i|}
\min \Bigg(
 \frac{\pi_\theta(o_{i,t}\mid q,o_{i,<t})}
       {\pi_{\theta_{\mathrm{old}}}(o_{i,t}\mid q,o_{i,<t})}
  \,\hat A_{i,t}, \\
& \mathrm{clip} \Bigg(
    \frac{\pi_\theta(o_{i,t}\mid q,o_{i,<t})}
         {\pi_{\theta_{\mathrm{old}}}(o_{i,t}\mid q,o_{i,<t})},
    \, 1-\varepsilon,\, 1+\varepsilon
  \Bigg)
  \hat A_{i,t}
\Bigg)
\end{aligned}
\label{eq:grpo}
\end{equation}
where $\theta$ and $\theta_{\mathrm{old}}$ denote the parameters of the current policy and the frozen reference (behavior) policy used to generate samples, respectively. 
For a given input prompt $q$, GRPO samples a group of $G$ output sequences $\{o_i\}_{i=1}^G$ from $\pi_{\theta_{\mathrm{old}}}$. 
Each $o_i = (o_{i,1}, \ldots, o_{i,|o_i|})$ is a token sequence with length $|o_i|$, and $o_{i,<t}$ denotes the prefix up to token $t-1$. 
$\pi_\theta(o_{i,t}\mid q,o_{i,<t})$ is the probability assigned by the current policy to token $o_{i,t}$ conditioned on $q$ and the prefix, and the ratio
$
\frac{\pi_\theta(o_{i,t}\mid q,o_{i,<t})}{\pi_{\theta_{\mathrm{old}}}(o_{i,t}\mid q,o_{i,<t})}
$
is the importance-sampling weight. 
$\hat{A}_{i,t}$ denotes the (estimated) advantage associated with token $t$ in sequence $o_i$, computed from the scalar rewards by normalizing performance within the sampled group. 
$\varepsilon$ is the clipping hyperparameter, and $\mathrm{clip}(\cdot, 1-\varepsilon, 1+\varepsilon)$ truncates the ratio to stabilize updates. 
The objective averages the clipped surrogate loss over tokens and over the $G$ sampled generations.


\subsection{AI detection and localization}
For image tampering detection, different studies have had various approaches. One study used supervised finetuning to provide explainable predictions using a VLM \cite{fakeshield}. Multimodal LLM framework for image forgery detection and localization was done by combining pixel-level mask extraction with explainable, interactive dialogue\cite{forgegpt}. Another study performs comprehensive clue fusion via a Chain-of-Clues prompt, combining multiple visual and textual cues to generate segmentation maps that pinpoint tampered regions\cite{forgerysleuth}. Another framework detects and localizes image deepfakes by predicting tampering masks and generating textual explanations, trained on the large, diverse SID-Set dataset\cite{sida}. Recent GRPO/RL-based methods \cite{thinkfake,fakehr1,ivyfake,raidx} primarily target whole-image real/fake AIGC classification. In contrast, our setting addresses locally edited images, requiring both tamper detection and spatial localization. This motivates our localization-aware GRPO reward, where edited images are rewarded using bounding-box IoU rather than only binary correctness. The VLM output is also used functionally by passing its reasoning trace and bounding box to the segmentation module. Thus, while simple, our method provides a practical decoupled GRPO-based reasoning-and-localization framework for local edit detection, which has not been explored in prior work.

SFT-based explanation methods require stronger supervision, often using curated reasoning texts from third-party models. Our VLM learns reasoning from rewards without explicit explanation supervision.


\section{Methodology}

In this study, we develop and train an end-to-end framework for tampered image detection and localization.  As illustrated in Fig. \ref{fig:overall}, the tamper detection and localization modules are decoupled. We utilize the reasoning capabilities of Qwen2.5-VL trained via GRPO, a popular Reinforcement Learning (RL) technique, as the VLM. Then, another text-guided segmentation model is used for localization and for generating pixel-level binary masks. The core functionality of this model is tied to the carefully designed reward functions that guide the model toward meaningful reasoning and accurate prediction. 

Training is performed in two stages. First, GRPO optimizes the VLM to produce a tamper decision, reasoning trace, and coarse bounding box. Second, the VLM is frozen, and the segmentation module is trained using ground-truth tamper masks, with the VLM reasoning trace (examples given in Fig. \ref{fig:think_traces}) and predicted bbox serving as prompt inputs.

\subsection{Reward design}
\begin{figure}
    \centering
    \includegraphics[width=\linewidth]{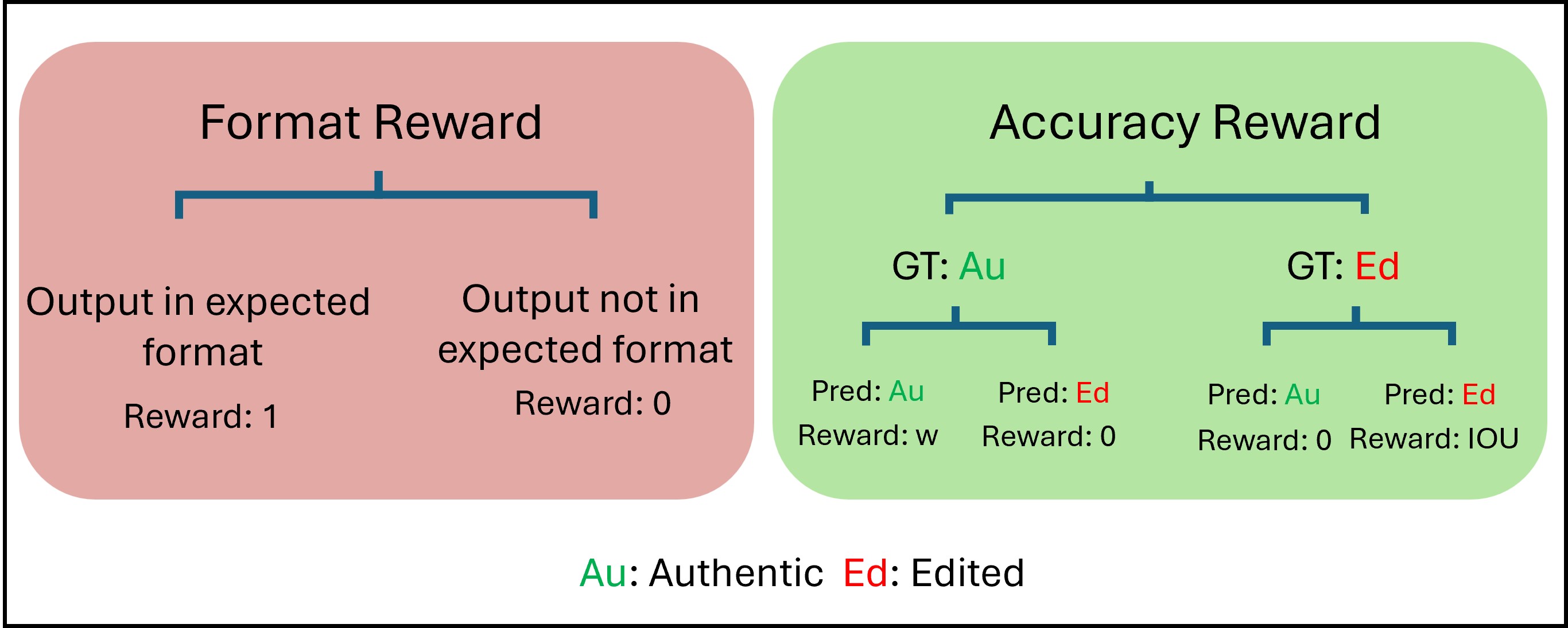}
    \caption{Reward design cases. 
    \textbf{Format reward (left):} The model receives a reward of 1 if the generated output follows the expected format, and 0 otherwise. 
    \textbf{Accuracy reward (right):} The reward depends on the relationship between the ground-truth label and the model prediction. 
    For authentic images correctly predicted as authentic, a fixed reward $w$ is assigned. 
    For edited images correctly predicted as edited, the reward is the IoU between the predicted and ground-truth bounding boxes. 
    Incorrect predictions receive zero reward.}
    \label{fig:rewards}
\end{figure}

The crucial part of this study was the reward design as it basically determines the model's performance and its reasoning capabilities. For simplicity, we employed only two reward functions. Accuracy reward and format reward proved to be strong and simple reward functions for the model to learn. Different reward cases are illustrated in the Fig. \ref{fig:rewards}.

\subsubsection{Format Reward Function}
Format reward is applied for generating outputs in the expected format. It is crucial as we expect the model to provide the bounding box pixel coordinates following a reasoning trace. The expected format is 
\par\noindent
\resizebox{\linewidth}{!}{%
\textbf{\texttt{\textless think\textgreater ...\textless /think\textgreater\textless answer\textgreater [...]\textless /answer\textgreater}}%
}
\par\noindent
If the format is exact, we assign a reward of 1 and 0 if incorrect as depicted in Fig. \ref{fig:rewards}. Additionally, any generation with an incorrect format is not considered for the accuracy reward. The format reward equation is given by Eq. \ref{eq:format}, where $y, \mathcal{F} \text{ and } \mathbb{I}[.]$ represent the model generation, correct format space and indicator function respectively. 

\begin{equation}
    R_{fmt}(y) = \mathbb{I}[y\in \mathcal{F}]
    \label{eq:format}
\end{equation}

\subsubsection{Accuracy Reward Function}
This measures how accurate a generation from the model is. As we are using both tampered and non-tampered inputs, we have 4 cases to consider as shown in Fig. \ref{fig:rewards}. For the two incorrect prediction cases, the accuracy reward is 0. The true-negative (authentic image predicted as authentic) contributes to a reward value of $w$, which is a hyperparameter to be tuned. For the true-positive case (edited image predicted as edited), we have to compute how close the prediction is to the ground truth. We compare the localization prediction area to the ground truth edit area. For that, we use intersection over union (IoU) as a metric. It compares the overlapping region as a ratio to the total area of ground truth and predicted areas. It is given in Eq. \ref{eq:iou}, where $A, B$ are predicted and ground truth edit areas respectively.

\begin{equation}
\mathrm{IoU}(A, B) = \frac{|A \cap B|}{|A \cup B|}
\label{eq:iou}
\end{equation}

The accuracy reward as a whole is represented in Eq. \ref{eq:acc} where $y, \mathcal{F},  \mathbb{I}[.], B, L \text{ and } w$ represent the model generation, correct format space, indicator function, bounding box, label and real-prediction reward value. $w$ is a hyperparameter. 

\begin{equation}
\begin{aligned}
R_{\text{acc}}(y)
= &\mathbb{I}[y \in \mathcal{F}] \cdot 
\\&\Big(
 \text{IoU}(B_{gt}, B_{pred}) 
  \cdot \mathbb{I}[L_{gt} = L_{pred} = \text{fake}] \\
& +\, w 
  \cdot \mathbb{I}[L_{gt} = L_{pred} = \text{real}]
\Big)
\end{aligned}
\label{eq:acc}
\end{equation}

Ablations on the true-negative reward weight and training reward curves are provided in the supp. material.


\subsection{GRPO-based VLM Training}

We train the vision-language model using Group Relative Policy Optimization (GRPO) to encourage explicit reasoning before making a tampering decision. Given an input image, the VLM generates a structured output consisting of a reasoning trace followed by a final answer indicating whether the image is tampered; for tampered images, the answer additionally includes a bounding box prediction.

For each training sample, we draw a group of candidate generations from the current policy. Each generation is evaluated using the reward functions described in the previous section, and relative advantages are computed within the group. These advantages are then used to update the VLM parameters via the GRPO objective.

Here, only the VLM parameters are optimized. The training focuses exclusively on improving reasoning quality and coarse localization behavior through reinforcement learning, without relying on explicit reasoning supervision.


\subsection{Forgery Segmentation}

The vision-language model (VLM) produces a bounding box as its final prediction, which only provides a coarse spatial localization and does not directly yield pixel-level delineation of the manipulated region. To obtain fine-grained segmentation masks, we introduce a prompt-conditioned segmentation stage that refines the VLM output into a dense prediction. This stage leverages both the full reasoning trace generated by the VLM and the predicted bounding box coordinates to guide the segmentation process.

By separating coarse localization from pixel-wise segmentation, the VLM is unburdened from learning dense spatial supervision and can focus on semantic reasoning and tampering identification. The segmentation module, in turn, specializes in spatial refinement using the high-level cues produced by the VLM. This modular design is inspired by recent work on prompt-driven segmentation using multimodal reasoning signals \cite{fakeshield}, but is adapted to operate on reasoning traces and explicit bounding box outputs.

\paragraph{Reasoning-Guided Prompt Encoding.}
Let $I$ denote the input image and $R$ the complete output of the VLM, consisting of the reasoning trace and predicted bounding box $B$. We encode $(I, R)$ using a multimodal encoder $\Phi(\cdot)$ that jointly processes visual and textual inputs. A special segmentation token $\langle \mathrm{SEG} \rangle$ is appended to the text input, and its final hidden representation
\begin{equation}
\mathbf{z}_{\mathrm{seg}} = \Phi(I, R)_{\langle \mathrm{SEG} \rangle}
\end{equation}
serves as a compact embedding of the reasoning and localization cues. This embedding is mapped to a segmentation prompt via a learnable projection
\begin{equation}
\mathbf{q} = f_{\mathrm{proj}}(\mathbf{z}_{\mathrm{seg}}),
\end{equation}
where $f_{\mathrm{proj}}(\cdot)$ is a lightweight neural network trained jointly with the encoder.

\paragraph{Segmentation with SAM.}
We adopt the Segment Anything Model (SAM) \cite{sam} as the segmentation backbone and keep all its parameters frozen. The image is encoded as visual features
\begin{equation}
\mathbf{v} = E_{\mathrm{SAM}}(I),
\end{equation}
and the final pixel-level mask $\hat{M}$ is predicted by conditioning SAM’s mask decoder on the image features, the projected prompt, and the bounding box:
\begin{equation}
\hat{M} = D_{\mathrm{SAM}}(\mathbf{v}, \mathbf{q}, B).
\end{equation}
Here, the bounding box provides a coarse spatial prior, while the prompt embedding injects high-level semantic information derived from the VLM reasoning trace.

\paragraph{Training Objective.}
Given a ground-truth tampering mask $M$, the segmentation module is trained using a combination of pixel-wise binary cross-entropy and region-based overlap loss:
\begin{equation}
\mathcal{L}_{\text{pix}} = - \sum_i \left[ M_i \log \hat{M}_i + (1 - M_i)\log(1 - \hat{M}_i) \right],
\end{equation}
\begin{equation}
\mathcal{L}_{\text{overlap}} = 1 - \frac{2 \sum_i \hat{M}_i M_i}{\sum_i \hat{M}_i + \sum_i M_i + \epsilon}.
\end{equation}
The final loss is defined as
\begin{equation}
\mathcal{L}_{\text{seg}} = \mathcal{L}_{\text{pix}} + \lambda \cdot  \mathcal{L}_{\text{overlap}},
\end{equation}
where $\lambda$ balances the contribution of the two terms.

\paragraph{Optimization Details.}
During training, only the reasoning-guided prompt encoder and the projection module are updated. The segmentation backbone remains fixed. This design enforces that learning is concentrated on translating high-level reasoning and coarse localization into effective segmentation prompts, while preserving the generalization capability of the pretrained segmentation model.

\section{Experiments}
\subsection{Dataset and Experiment Setup}
Qwen2.5-VL-7B-Instruct\cite{qwen2.5-VL} was selected as the main model for experiments. The Qwen2.5-VL-7B model is relatively smaller while being powerful which makes it suitable for experimentation. We also tried different experiments with Qwen2.5-VL-3B\cite{qwen2.5-VL} and Gemma 3 4B\cite{gemma_2025} model for comparison. The training datasets consisted of AutoSplice\cite{autosplice},  CASIAv2\cite{casiav2}, Fantastic Reality\cite{fantastic_reality}, FFHQ-FM\cite{ffhqfm}, MagicBrush\cite{magicbrush} and SD\_inpaint\cite{fakeshield} datasets. These datasets included a wide array of image tampering types such as photoshops, diffusion inpainting and DeepFakes. 

\subsection{Metrics}

Since our model makes predictions for both tampered and authentic images, we evaluate it on (1) tamper detection and (2) localization quality. Detection performance is measured using classification accuracy (Acc), while localization performance is measured using mean Intersection over Union (mIoU).

We further introduce a unified metric, \textbf{effective Intersection over Union (eff-IoU)}, defined as:
\begin{equation}
\text{eff-IoU} = \text{Acc} \times \text{mIoU}, \quad \text{eff-IoU} \in [0,1].
\end{equation}

This metric captures both detection and localization performance in a single value. We additionally report Pixel-F1 and discuss its relationship to eff-IoU in the supplementary material.

Evaluation is conducted on the test splits of AutoSplice\cite{autosplice}, MagicBrush\cite{magicbrush}, FFHQ-FM\cite{ffhqfm}, and SD\_inpaint\cite{fakeshield}, with balanced numbers of tampered and authentic images to ensure fairness. Additional dataset-overlap discussion is provided in the supplementary material.

\begin{table*}[t]
\centering
\footnotesize
\resizebox{\textwidth}{!}{
\begin{tabular}{l ccc ccc ccc ccc}
\toprule
Model 
& \multicolumn{3}{c}{AutoSplice}
& \multicolumn{3}{c}{MagicBrush}
& \multicolumn{3}{c}{FFHQ-FM}
& \multicolumn{3}{c}{SD\_inpaint} \\
& Acc$\uparrow$ & mIoU$\uparrow$ & eff-IoU$\uparrow$
& Acc$\uparrow$ & mIoU$\uparrow$ & eff-IoU$\uparrow$
& Acc$\uparrow$ & mIoU$\uparrow$ & eff-IoU$\uparrow$
& Acc$\uparrow$ & mIoU$\uparrow$ & eff-IoU$\uparrow$ \\
\midrule
\shortstack[c]{Qwen2.5-VL(vanilla)}
& 0.567 & 0.028 & 0.016
& 0.545 & 0.017 & 0.009
& 0.536 & 0.001 & 0.000
& 0.524 & 0.006 & 0.003 \\

\shortstack{AdaIFL [2025]\cite{adaifl}} 
& 0.537 & 0.267 & 0.143
& 0.543 & 0.120 & 0.065
& 0.472 & 0.172 & 0.081
& 0.486 & 0.069 & 0.033 \\

\shortstack{Mesorch [2025]\cite{mesorch}}
& 0.500 & 0.269 & 0.135
& 0.500 & 0.137 & 0.068
& 0.465 & 0.034 & 0.016
& 0.500 & 0.083 & 0.042 \\

FakeShield [2025]\cite{fakeshield} 
& 0.537 & \cellcolor{second}0.497 & 0.267
& 0.507 & \cellcolor{second}0.151 & 0.076
& 0.479 & \cellcolor{second}0.243 & \cellcolor{second}0.116
& 0.504 & \cellcolor{best}{0.166} & \cellcolor{second}0.084 \\

\shortstack{SIDA [2025]\cite{sida}}
& \cellcolor{second}0.609 & 0.467 & \cellcolor{second}0.284
& \cellcolor{second}0.808 & \cellcolor{best}0.162 & \cellcolor{second}0.131
& \cellcolor{second}0.554 & 0.190 & 0.105
& \cellcolor{second}0.627 & 0.090 & 0.056 \\


\shortstack{Ours} 
& \cellcolor{best}0.842 & \cellcolor{best}{0.513} & \cellcolor{best}{0.432}
& \cellcolor{best}{0.951} & 0.141 & \cellcolor{best}0.134
& \cellcolor{best}{0.590} & \cellcolor{best}0.247 & \cellcolor{best}0.146
& \cellcolor{best}{0.850} & \cellcolor{second}0.155 & \cellcolor{best}{0.132} \\
\bottomrule
\end{tabular}
}
\caption{Performance comparison across datasets. 
Accuracy (Acc), mean Intersection-over-Union (mIoU), and effective IoU (eff-IoU) are reported for each method. 
The effective IoU is defined as \textbf{$\text{eff-IoU} = \text{Acc} \times \text{mIoU}$} to jointly reflect classification and localization performance. 
The \colorbox{best}{best} and \colorbox{second}{second-best} results for each metric are highlighted.}
\label{tab:performance_comparison_transposed}
\end{table*}

\begin{table}[t]
\centering
\footnotesize
\begin{tabular}{l c c c c}
\toprule
Rank & Model & Avg Acc$\uparrow$ & Avg mIoU$\uparrow$ & Avg eff-IoU$\uparrow$ \\
\midrule
1 & \shortstack{Ours} & \cellcolor{best}{0.808} & \cellcolor{best}0.264 & \cellcolor{best}{0.211} \\
2 & \shortstack{SIDA [2025]\cite{sida}} & \cellcolor{second}0.650 & 0.227 & \cellcolor{second}0.144 \\
3 & FakeShield [2025]\cite{fakeshield} & 0.507 & \cellcolor{second}0.263 & 0.136 \\
4 & \shortstack{AdaIFL [2025]\cite{adaifl}} & 0.510 & 0.157 & 0.081 \\
5 & \shortstack{Mesorch [2025]\cite{mesorch}} & 0.491 & 0.131 & 0.065 \\
6 & \shortstack[c]{Qwen2.5-VL(vanilla)} & 0.543 & 0.013 & 0.007 \\
\bottomrule
\end{tabular}
\caption{Models ranking based on the effective IoU (eff-IoU) averaged across datasets. 
Average metrics are reported. 
The \colorbox{best}{best} and \colorbox{second}{second-best} values for each metric are highlighted.}
\label{tab:avg_performance_ranked}
\end{table}

\begin{table}[t]
\centering
\footnotesize
\begin{tabular}{l c}
\toprule
Model & Avg eff-IoU$\uparrow$ \\
\midrule
Qwen2.5-VL-7B\cite{qwen2.5-VL} & 0.1381 \\
Qwen2.5-VL-3B\cite{qwen2.5-VL} & 0.1359 \\
Gemma 3 4B\cite{gemma_2025} & 0.1342 \\
\bottomrule
\end{tabular}
\caption{Comparison of different VLM backbones.
Average effective IoU (eff-IoU) across all datasets is reported for each model. 
All models, including the Qwen2.5-VL-7B, which we use as our final model, are trained for a single epoch under the same conditions for fair comparison.}
\label{tab:vlm_backbone_comparison}
\end{table}

\begin{figure*}[b!]
    \centering
    \includegraphics[width=\textwidth]{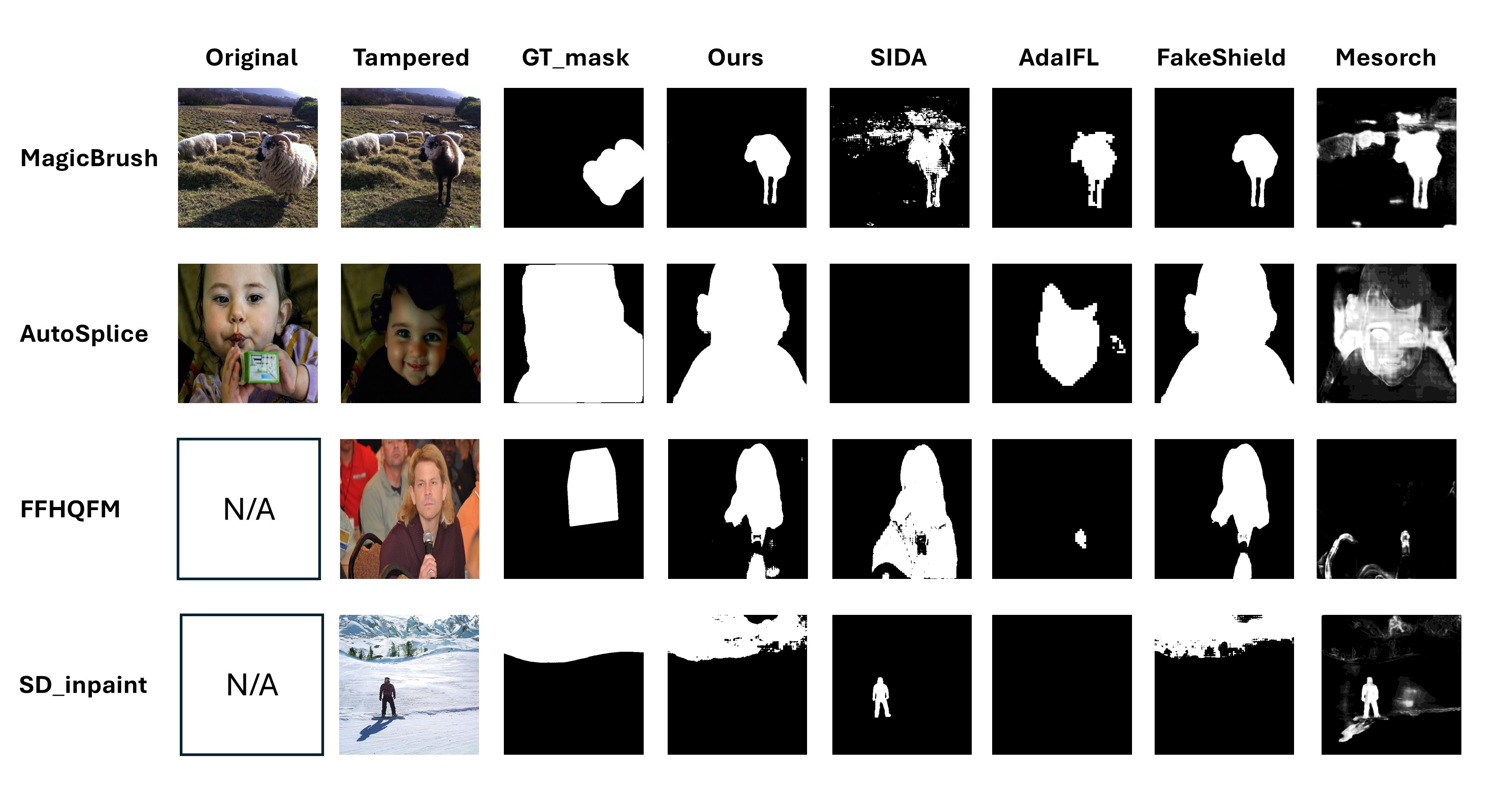}
    \caption{Qualitative comparison of tampering localization results across datasets. 
    Each row corresponds to a dataset, and columns show the original image, tampered image, ground-truth mask, and predicted masks from different methods. 
    For the FFHQ-FM and SD\_inpaint datasets, paired authentic images are not provided.}
    \label{fig:qualitative}
\end{figure*}

\begin{figure*}[t]
    \centering
    \includegraphics[width=\textwidth]{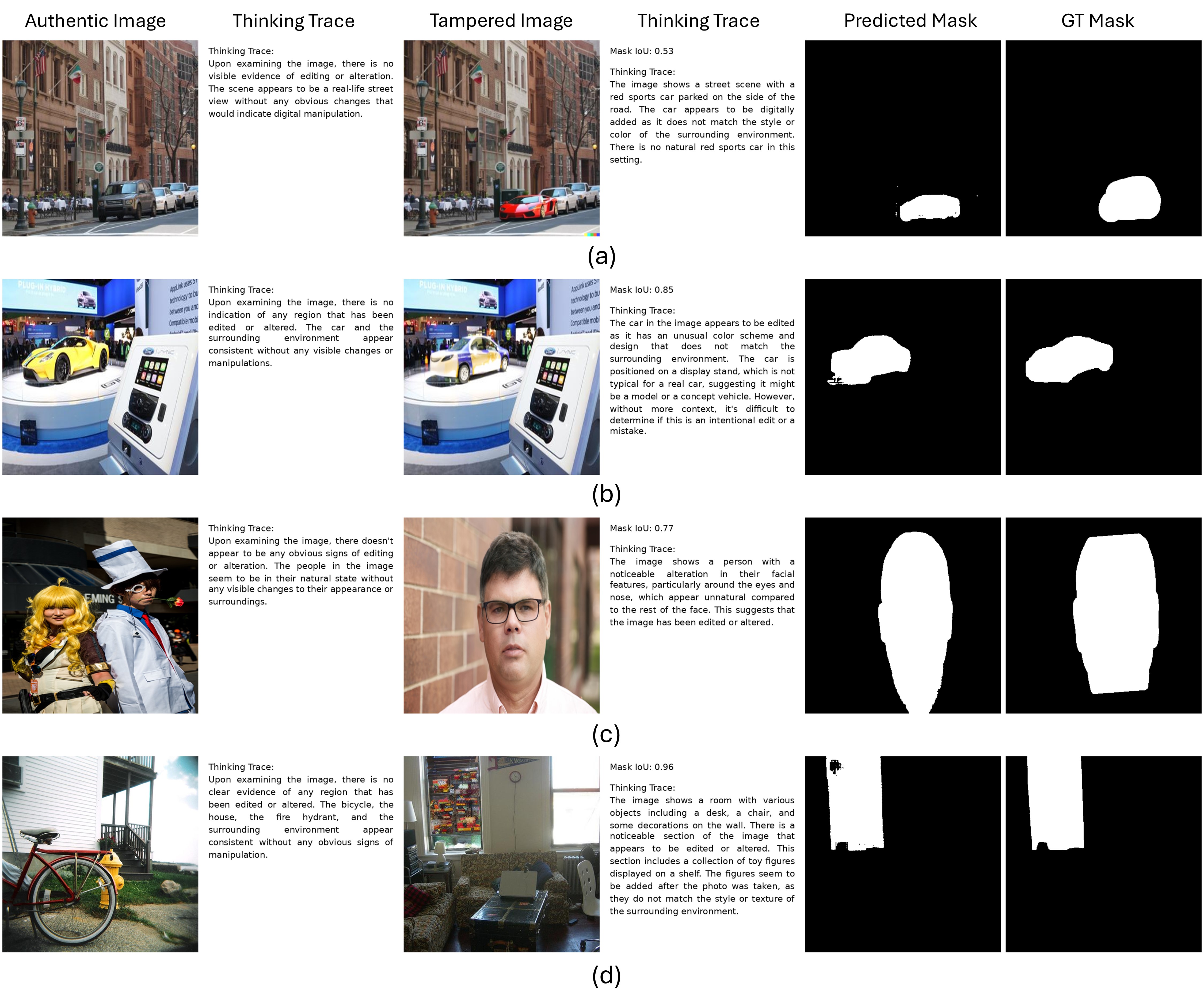}
    \caption{Visualization of model predictions and reasoning traces on multiple datasets. 
    From left to right: authentic image, reasoning trace for the authentic image, corresponding 
    tampered image, reasoning trace for the tampered image, predicted mask, and ground-truth mask. The rows are (a) MagicBrush, (b) Autosplice, (c) FFHQ-FM, (d) SD\_inpaint. FFHQ-FM and SD\_inpaint do not provide one-to-one authentic–tampered image pairs, the authentic 
    and tampered examples shown are selected independently.}
    \label{fig:think_traces}
\end{figure*}

\subsection{Baselines}
We compare our method against several representative image manipulation localization methods such as \textbf{Mesorch \cite{mesorch}}, \textbf{AdaIFL \cite{adaifl}}, \textbf{FakeShield \cite{fakeshield}}, \textbf{SIDA \cite{sida}}.

\subsection{Results}

The three quantitative evaluation metrics are: detection accuracy (Acc), mean intersection-over-union (mIoU), and effective IoU (eff-IoU). As shown in Table~\ref{tab:performance_comparison_transposed}, our method achieves the highest detection accuracy across all datasets. In terms of localization performance (mIoU), our approach remains competitive and consistently ranks among the top two methods. The combined metric eff-IoU further demonstrates the effectiveness of our approach by achieving the best eff-IoU for all datasets, indicating strong joint performance for detection and localization. Table \ref{tab:avg_performance_ranked} reports the average performance across datasets and ranks the models based on eff-IoU. Our method achieves the highest average eff-IoU, outperforming the second-ranked method by more than 46\%.

Fig.\ref{fig:qualitative} presents a qualitative comparison of predictions. Compared to baselines, we produce better spatially coherent masks and contain fewer spurious activations while aligning closely with the GT tampered areas. Some baselines produce fragmented masks or introduce false positives outside the manipulated regions. 

Fig. \ref{fig:think_traces} demonstrates the reasoning traces generated by our model on both authentic and tampered inputs. These examples show that the GRPO-trained model often reasons over forensic cues. In several paired authentic/edited examples, the model correctly distinguishes visually similar images that may deceive the naked eye.

\subsection{Discussion}

Based on our findings, we show promising results to leverage GRPO and weak supervision to train a VLM to reason about AI edits with competitive performance. We want to discuss two potential directions related to scalability:


1. \textbf{Model Backbone:} 
We evaluated the proposed framework across VLM backbones of different sizes (trained under same conditions). As shown in Table ~\ref{tab:vlm_backbone_comparison}, performance improves slightly with model capacity, but the differences remain small, suggesting that reward design, training data, and training duration may be more influential than backbone size in this setting. Due to computational constraints, we leave evaluation with larger VLMs for future work.


2. \textbf{Data:} With a fixed VLM backbone, performance may be improved by scaling the training data. However, this requires both image-level labels and ground-truth edit masks of AI edits, which remain limited in public datasets. These limitations suggest that investing in higher-quality data and annotations is an important and worthwhile direction for scaling performance. Further analysis of input image noise is provided in the supplementary material.

\vspace*{1in}

\section{Conclusion}
In this work, we studied whether vision-language models can be trained to reason about AI-edited images under weak supervision. We proposed a GRPO-based training framework that encourages the model to produce an explicit reasoning trace before predicting tampering, using only simple format and accuracy rewards. To obtain pixel-level localization, we decouple detection from segmentation by refining the VLM’s coarse bounding-box prediction into a dense mask with a prompt-conditioned segmentation module. 

Experiments across multiple manipulation datasets show that our approach achieves strong detection accuracy and competitive localization performance by leveraging inherent reasoning capabilities of VLMs compared to state-of-the-art baselines, while requiring substantially less supervision than explanation-based supervised finetuning methods. Qualitative examples further suggest that the trained model can produce reasoning traces grounded in forensic cues, and we introduce effective-IoU as a unified metric to jointly assess detection and localization quality.

Future work will explore scaling to larger VLM backbones and broader training data, as well as improving robustness to distribution shifts and challenging perturbations, to further strengthen reasoning-driven forensic localization.

\section{Acknowledgment}
This work was supported by IBM through the IBM-Rensselaer Future of Computing Research Collaboration.

{
    \small
    \bibliographystyle{ieeenat_fullname}
    \bibliography{main}

@String(CVPR  = {IEEE Conf. Comput. Vis. Pattern Recog.})

@String(ICCV  = {Int. Conf. Comput. Vis.})

@String(NeurIPS = {Adv. Neural Inform. Process. Syst.})

@String(AAAI  = {AAAI})

@String(CVPR  = {CVPR})

@String(ICCV  = {ICCV})

@String(NeurIPS = {NeurIPS})

@INPROCEEDINGS{ImageManipulationDetection,
  author={Chen, Xinru and Dong, Chengbo and Ji, Jiaqi and Cao, Juan and Li, Xirong},
  booktitle={2021 IEEE/CVF International Conference on Computer Vision (ICCV)}, 
  title={Image Manipulation Detection by Multi-View Multi-Scale Supervision}, 
  year={2021},
  volume={},
  number={},
  pages={14165-14173},
  doi={10.1109/ICCV48922.2021.01392}}

@INPROCEEDINGS{HierarchicalFine-Grained,
  author={Guo, Xiao and Liu, Xiaohong and Ren, Zhiyuan and Grosz, Steven and Masi, Iacopo and Liu, Xiaoming},
  booktitle={2023 IEEE/CVF Conference on Computer Vision and Pattern Recognition (CVPR)}, 
  title={Hierarchical Fine-Grained Image Forgery Detection and Localization}, 
  year={2023},
  volume={},
  number={},
  pages={3155-3165},
  doi={10.1109/CVPR52729.2023.00308}}

@misc{guo_deepseek-r1_2025,
      title={DeepSeek-R1: Incentivizing Reasoning Capability in LLMs via Reinforcement Learning}, 
      author={{DeepSeek-AI}},
      year={2025},
      eprint={2501.12948},
      archivePrefix={arXiv},
      primaryClass={cs.CL},
      url={https://arxiv.org/abs/2501.12948}, 
}

@misc{forgegpt,
      title={ForgeryGPT: Multimodal Large Language Model For Explainable Image Forgery Detection and Localization}, 
      author={Jiawei Liu and Fanrui Zhang and Jiaying Zhu and Esther Sun and Qiang Zhang and Zheng-Jun Zha},
      year={2025},
      eprint={2410.10238},
      archivePrefix={arXiv},
      primaryClass={cs.CV},
      url={https://arxiv.org/abs/2410.10238}, 
}

@article{forgerysleuth,
  title={ForgerySleuth: Empowering Multimodal Large Language Models for Image Manipulation Detection},
  author={Zhihao Sun and Haoran Jiang and Haoran Chen and Yixin Cao and Xipeng Qiu and Zuxuan Wu and Yu-Gang Jiang},
  journal={ArXiv},
  year={2024},
  volume={abs/2411.19466},
  url={https://api.semanticscholar.org/CorpusID:274422867}
}

@misc{sam,
      title={Segment Anything}, 
      author={Alexander Kirillov and Eric Mintun and Nikhila Ravi and Hanzi Mao and Chloe Rolland and Laura Gustafson and Tete Xiao and Spencer Whitehead and Alexander C. Berg and Wan-Yen Lo and Piotr Dollár and Ross Girshick},
      year={2023},
      eprint={2304.02643},
      archivePrefix={arXiv},
      primaryClass={cs.CV},
      url={https://arxiv.org/abs/2304.02643}, 
}

@inproceedings{mesorch,
  title={Mesoscopic insights: orchestrating multi-scale \& hybrid architecture for image manipulation localization},
  author={Zhu, Xuekang and Ma, Xiaochen and Su, Lei and Jiang, Zhuohang and Du, Bo and Wang, Xiwen and Lei, Zeyu and Feng, Wentao and Pun, Chi-Man and Zhou, Ji-Zhe},
  booktitle={Proceedings of the AAAI Conference on Artificial Intelligence},
  volume={39},
  number={10},
  pages={11022--11030},
  year={2025}
}

@inproceedings{fakeshield,
        title={FakeShield: Explainable Image Forgery Detection and Localization via Multi-modal Large Language Models},
        author={Xu, Zhipei and Zhang, Xuanyu and Li, Runyi and Tang, Zecheng and Huang, Qing and Zhang, Jian},
        booktitle={International Conference on Learning Representations},
        year={2025}
}

@inproceedings{sida,
  author       = {Zhenglin Huang and Jinwei Hu and Xiangtai Li and Yiwei He and Xingyu Zhao and Bei Peng and Baoyuan Wu and Xiaowei Huang and Guangliang Cheng},
  title        = {{SIDA:} Social Media Image Deepfake Detection, Localization and Explanation
                  with Large Multimodal Model},
  booktitle    = {{IEEE/CVF} Conference on Computer Vision and Pattern Recognition(CVPR) 2025},
  year         = {2025},
}

@inproceedings{adaifl,
  title={AdaIFL: Adaptive Image Forgery Localization via a Dynamic and Importance-Aware Transformer Network},
  author={Li, Yuxi and Cheng, Fuyuan and Yu, Wangbo and Wang, Guangshuo and Luo, Guibo and Zhu, Yuesheng},
  booktitle={European Conference on Computer Vision},
  pages={477--493},
  year={2025},
  organization={Springer}
}

@inproceedings{autosplice,
  title={AutoSplice: A Text-prompt Manipulated Image Dataset for Media Forensics},
  author={Jia, Shan and Huang, Mingzhen and Zhou, Zhou and Ju, Yan and Cai, Jialing and Lyu, Siwei},
  booktitle={Proceedings of the IEEE/CVF Conference on Computer Vision and Pattern Recognition},
  pages={893--903},
  year={2023}
}

@inproceedings{magicbrush,
        title={MagicBrush: A Manually Annotated Dataset for Instruction-Guided Image Editing},
        author={Kai Zhang and Lingbo Mo and Wenhu Chen and Huan Sun and Yu Su},
        booktitle={Advances in Neural Information Processing Systems},
        year={2023}
}

@misc{ffhqfm,
  doi = {10.48550/ARXIV.2208.11776},
  url = {https://arxiv.org/abs/2208.11776},
  author = {DeCann, Brian and Trapeznikov, Kirill},
  title = {Comprehensive Dataset of Face Manipulations for Development and Evaluation of Forensic Tools},
  publisher = {arXiv},
  year = {2022},
  copyright = {Creative Commons Attribution 4.0 International}
}

@INPROCEEDINGS{casiav2,
  author={Dong, Jing and Wang, Wei and Tan, Tieniu},
  booktitle={2013 IEEE China Summit and International Conference on Signal and Information Processing}, 
  title={CASIA Image Tampering Detection Evaluation Database}, 
  year={2013},
  volume={},
  number={},
  pages={422-426},
  doi={10.1109/ChinaSIP.2013.6625374}}

@inproceedings{fantastic_reality,
  title={The Point Where Reality Meets Fantasy: Mixed Adversarial Generators for Image Splice Detection},
  author={Kniaz, Vladimir V. and Knyaz, Vladimir A. and Remondino, Fabio},
  booktitle={Advances in Neural Information Processing Systems (NeurIPS)},
  volume={32},
  pages={215--226},
  year={2019}
}

@misc{qwen2.5-VL,
    title = {Qwen2.5-VL},
    url = {https://qwenlm.github.io/blog/qwen2.5-vl/},
    author = {{Qwen Team}},
    month = {January},
    year = {2025}
}

@misc{gemma_2025,
    title={Gemma 3},
    url={https://goo.gle/Gemma3Report},
    publisher={Kaggle},
    author={{Gemma Team}},
    year={2025}
}

@inproceedings{gpt3,
 author = {Brown, Tom and Mann, Benjamin and Ryder, Nick and Subbiah, Melanie and Kaplan, Jared D and Dhariwal, Prafulla and Neelakantan, Arvind and Shyam, Pranav and Sastry, Girish and Askell, Amanda and Agarwal, Sandhini and Herbert-Voss, Ariel and Krueger, Gretchen and Henighan, Tom and Child, Rewon and Ramesh, Aditya and Ziegler, Daniel and Wu, Jeffrey and Winter, Clemens and Hesse, Chris and Chen, Mark and Sigler, Eric and Litwin, Mateusz and Gray, Scott and Chess, Benjamin and Clark, Jack and Berner, Christopher and McCandlish, Sam and Radford, Alec and Sutskever, Ilya and Amodei, Dario},
 booktitle = {Advances in Neural Information Processing Systems},
 editor = {H. Larochelle and M. Ranzato and R. Hadsell and M.F. Balcan and H. Lin},
 pages = {1877--1901},
 publisher = {Curran Associates, Inc.},
 title = {Language Models are Few-Shot Learners},
 url = {https://proceedings.neurips.cc/paper_files/paper/2020/file/1457c0d6bfcb4967418bfb8ac142f64a-Paper.pdf},
 volume = {33},
 year = {2020}
}

@article{gemini,
author = {Anil, Rohan and Borgeaud, Sebastian and Alayrac, Jean-Baptiste and Yu, Jiahui and Soricut, Radu and Schalkwyk, Johan and Dai, Andrew and Hauth, Anja and Millican, Katie and Johnson, Melvin and Antonoglou, Ioannis and Schrittwieser, Julian and Glaese, Amelia and Chen, Jilin and Pitler, Emily and Lillicrap, Timothy and Lazaridou, Angeliki and Firat, Orhan and Vinyals, Oriol},
year = {2023},
month = {12},
pages = {},
title = {Gemini: A Family of Highly Capable Multimodal Models},
doi = {10.48550/arXiv.2312.11805}
}

@InProceedings{diffusion_1,
  title = 	 {Improved Denoising Diffusion Probabilistic Models},
  author =       {Nichol, Alexander Quinn and Dhariwal, Prafulla},
  booktitle = 	 {Proceedings of the 38th International Conference on Machine Learning},
  pages = 	 {8162--8171},
  year = 	 {2021},
  editor = 	 {Meila, Marina and Zhang, Tong},
  volume = 	 {139},
  series = 	 {Proceedings of Machine Learning Research},
  month = 	 {18--24 Jul},
  publisher =    {PMLR},
}

@misc{diffusion_2,
      title={High-Resolution Image Synthesis with Latent Diffusion Models}, 
      author={Robin Rombach and Andreas Blattmann and Dominik Lorenz and Patrick Esser and Björn Ommer},
      year={2021},
      eprint={2112.10752},
      archivePrefix={arXiv},
      primaryClass={cs.CV}
}

@article{forgery_det_1,
title = {A survey on deep learning-based image forgery detection},
journal = {Pattern Recognition},
volume = {144},
pages = {109778},
year = {2023},
issn = {0031-3203},
doi = {10.1016/j.patcog.2023.109778},
url = {https://www.sciencedirect.com/science/article/pii/S0031320323004764},
author = {Fatemeh Zare Mehrjardi and Ali Mohammad Latif and Mohsen Sardari Zarchi and Razieh Sheikhpour},
}

@ARTICLE{forgery_det_2,
  author={Wang, Junke and Li, Zhenxin and Zhang, Chao and Chen, Jingjing and Wu, Zuxuan and Davis, Larry S. and Jiang, Yu-Gang},
  journal={Proceedings of the IEEE}, 
  title={Fighting Malicious Media Data: A Survey on Tampering Detection and Deepfake Detection}, 
  year={2025},
  volume={113},
  number={3},
  pages={287-311},
  doi={10.1109/JPROC.2025.3576367}}

@ARTICLE{metadata_1,
  author={Lukas, J. and Fridrich, J. and Goljan, M.},
  journal={IEEE Transactions on Information Forensics and Security}, 
  title={Digital camera identification from sensor pattern noise}, 
  year={2006},
  volume={1},
  number={2},
  pages={205-214},
  doi={10.1109/TIFS.2006.873602}}

@article{metadata_2,
  title={Noiseprint: A CNN-Based Camera Model Fingerprint},
  author={D. Cozzolino and L. Verdoliva},
  journal={IEEE Transactions on Information Forensics and Security},
  doi={10.1109/TIFS.2019.2916364},
  pages={144-159},
  year={2020},
  volume={15}
}

@inproceedings{hallu_1,
  title={Evaluating Object Hallucination in Large Vision-Language Models},
  author={Li, Yifan and Du, Yifan and Zhou, Kun and Wang, Jinpeng and Zhao, Wayne Xin and Wen, Ji-Rong},
  booktitle={Proceedings of the 2023 Conference on Empirical Methods in Natural Language Processing (EMNLP)},
  year={2023},
  url={https://openreview.net/forum?id=xozJw0kZXF}
}

@article{hallu_2,
  title={A survey on hallucination in large vision-language models},
  author={Liu, Hanchao and Xue, Wenyuan and Chen, Yifei and Chen, Dapeng and Zhao, Xiutian and Wang, Ke and Hou, Liping and Li, Rongjun and Peng, Wei},
  journal={arXiv preprint arXiv:2402.00253},
  year={2024}
}

@ARTICLE{watermark,
  author={Cox, I.J. and Kilian, J. and Leighton, F.T. and Shamoon, T.},
  journal={IEEE Transactions on Image Processing}, 
  title={Secure spread spectrum watermarking for multimedia}, 
  year={1997},
  volume={6},
  number={12},
  pages={1673-1687},
  doi={10.1109/83.650120}}

@misc{spectrum,
    title={Watch your Up-Convolution: CNN Based Generative Deep Neural Networks are Failing to Reproduce Spectral Distributions},
    author={Ricard Durall and Margret Keuper and Janis Keuper},
    year={2020},
    eprint={2003.01826},
    archivePrefix={arXiv},
    primaryClass={cs.LG}
}

@inproceedings{lighting,
author = {Johnson, Micah and Farid, Hany},
year = {2007},
month = {06},
pages = {311-325},
title = {Exposing Digital Forgeries Through Specular Highlights on the Eye},
isbn = {978-3-540-77369-6},
journal = {Proceeding of the 9th International Workshop on Information Hiding},
doi = {10.1007/978-3-540-77370-2_21}
}

@article{thinkfake,
  publtype={informal},
  author={Tai-Ming Huang and Wei-Tung Lin and Kai-Lung Hua and Wen-Huang Cheng and Junichi Yamagishi and Jun-Cheng Chen},
  title={ThinkFake: Reasoning in Multimodal Large Language Models for AI-Generated Image Detection},
  year={2025},
  month={September},
  cdate={1756684800000},
  journal={CoRR},
  volume={abs/2509.19841},
  url={https://doi.org/10.48550/arXiv.2509.19841}
}

@article{ivyfake,
  title     = {Ivy-Fake: A Unified Explainable Framework and Benchmark for Image and Video AIGC Detection},
  author    = {Changjiang Jiang and Wenhui Dong and Zhonghao Zhang and Chenyang Si and Fengchang Yu and Wei Peng and Xinbin Yuan and Yifei Bi and Ming Zhao and Zian Zhou and Caifeng Shan},
  year      = {2025},
  url       = {https://arxiv.org/abs/2506.00979}
}

@inproceedings{raidx,
author = {Li, Tianxiao and Huang, Zhenglin and Wen, Haiquan and He, Yiwei and Lyu, Shuchang and Wu, Baoyuan and Cheng, Guangliang},
title = {RAIDX: A Retrieval-Augmented Generation and GRPO Reinforcement Learning Framework for Explainable Deepfake Detection},
year = {2025},
isbn = {9798400720352},
publisher = {Association for Computing Machinery},
address = {New York, NY, USA},
url = {https://doi.org/10.1145/3746027.3754798},
doi = {10.1145/3746027.3754798},

booktitle = {Proceedings of the 33rd ACM International Conference on Multimedia},
pages = {11746–11755},
numpages = {10},
location = {Dublin, Ireland},
series = {MM '25}
}

@unknown{fakehr1,
author = {Jiang, Changjiang and Sha, Xinkuan and Yu, Fengchang and Liu, Jingjing and Liu, Jian and Fang, Mingqi and Zhang, Chenfeng and Lu, Wei},
year = {2026},
month = {02},
pages = {},
title = {Fake-HR1: Rethinking Reasoning of Vision Language Model for Synthetic Image Detection},
doi = {10.48550/arXiv.2602.10042}
}
}

\clearpage
\appendix

\begin{center}
{\LARGE Supplementary Material}

\vspace{0.5em}

{\Large Can Vision-Language Models Reason about AI Edits in Images?}
\end{center}

\section{Reasoning Trace Analysis}
\label{sec:intro}

Without any supervision on the reasoning output, our model learns to leverage the model's inherent reasoning capabilities to make the decision on labeling and localizing the edited region. Some examples from each dataset clearly showcase this capability, as in Figures \ref{fig:trace_magicbrush}, \ref{fig:trace_autosplice}, \ref{fig:trace_ffhqfm}, \ref{fig:trace_sd_inpaint}.

\section{Ablation study on True-Negative Reward}
\begin{figure}[h]
  \centering

  \includegraphics[
    width=\linewidth,
    height=\rewardfigheight,
    trim=3pt 0pt 3pt 3pt,
    clip]{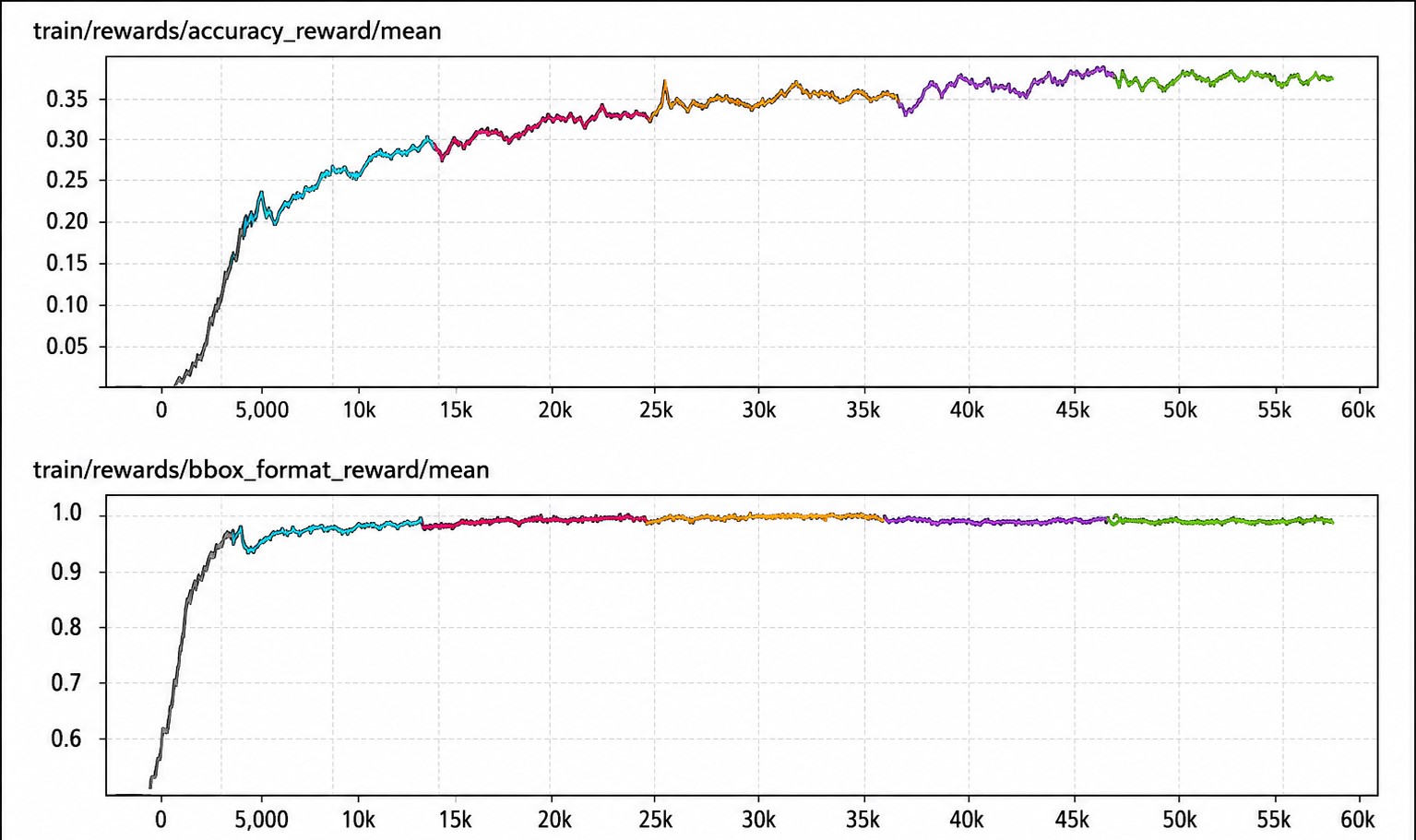}
  \vspace{-0.8em}

  \caption{Training reward curves (accuracy and format reward).}
  \label{fig:curves}
\end{figure}

As we mentioned in the main paper, the reward given for correctly predicting an authentic image as an untampered image was a hyperparameter. It had to be carefully tuned. Otherwise, the model would learn to hack the rewards by always predicting them as untampered to score on only the untampered data. Tab.~\ref{tab:tn_reward_ablation} studies the true-negative reward weight $w$ using only the VLM bounding-box output, before segmentation. The best setting balances authentic-image recognition and edited-region localization. The VLM alone already achieves high classification accuracy, while the segmentation module further improves fine localization later in the pipeline. The reward training curves are given in Fig. \ref{fig:curves}

\begin{table}[h]
\centering
\footnotesize
\setlength{\tabcolsep}{3pt}
\renewcommand{\arraystretch}{0.9}
\begin{tabular}{c c c c c c}
\toprule
TN Reward $w$ & 0.1 & \cellcolor{second}0.2 & 0.4 & 0.6 & 0.8 \\
\midrule
Classification Acc.$\uparrow$ & 0.850 & \cellcolor{second}0.959 & 0.534 & 0.604 & 0.500 \\
mIoU$\uparrow$ & 0.172 & \cellcolor{second}0.180 & 0.020 & 0.045 & 0.000 \\
\bottomrule
\end{tabular}
\caption{Effect of TN reward weight $w$ on model performance.}
\label{tab:tn_reward_ablation}
\end{table}

\section{eff-IoU and Pixel-F1}
We introduce eff-IoU as a novel metric in our study. The standard metric for measuring pixel-level detection is Pixel-F1. Therefore, we discuss the effectiveness of both metrics here. eff-IoU is not intended to replace Pixel-F1, but to complement it in our joint detection-localization setting. Pixel-F1 is standard for mask quality, but its treatment of authentic images depends on the empty-mask convention for real-real cases. Since our test set includes both authentic and tampered images, a practical model must first detect whether tampering exists and then localize it when present. eff-IoU makes this explicit by combining image-level accuracy with mIoU. Tab.~\ref{tab:performance_comparison_eff_pixel_f1} shows that Pixel-F1 and eff-IoU emphasize different behavior: on AutoSplice, SIDA has higher Pixel-F1, while our method has higher accuracy, mIoU, and eff-IoU, showing stronger joint detection-localization performance.

\begin{table*}[h]
\centering
\footnotesize
\resizebox{\textwidth}{!}{
\begin{tabular}{l cc cc cc cc}
\toprule
Model 
& \multicolumn{2}{c}{AutoSplice}
& \multicolumn{2}{c}{MagicBrush}
& \multicolumn{2}{c}{FFHQ-FM}
& \multicolumn{2}{c}{SD\_inpaint} \\
& eff-IoU$\uparrow$ & Pixel-F1$\uparrow$
& eff-IoU$\uparrow$ & Pixel-F1$\uparrow$
& eff-IoU$\uparrow$ & Pixel-F1$\uparrow$
& eff-IoU$\uparrow$ & Pixel-F1$\uparrow$ \\
\midrule

\shortstack{AdaIFL} 
& 0.143 & 0.259
& 0.065 & 0.209
& 0.081 & 0.161
& 0.033 & \cellcolor{second}0.178 \\

\shortstack{Mesorch}
& 0.135 & \cellcolor{second}0.350
& 0.068 & \cellcolor{second}0.210
& 0.016 & 0.139
& 0.042 & 0.150 \\

FakeShield
& 0.267 & 0.340
& 0.076 & 0.116
& \cellcolor{second}0.116 & 0.171
& \cellcolor{second}0.084 & 0.113 \\

\shortstack{SIDA}
& \cellcolor{second}0.284 & \cellcolor{best}0.484
& \cellcolor{second}0.131 & \cellcolor{best}0.436
& 0.105 & \cellcolor{best}0.356
& 0.056 & \cellcolor{best}0.381 \\

\shortstack{Ours} 
& \cellcolor{best}{0.432} & 0.340
& \cellcolor{best}{0.134} & 0.157
& \cellcolor{best}{0.146} & \cellcolor{second}0.183
& \cellcolor{best}{0.132} & 0.143 \\
\bottomrule
\end{tabular}
}
\caption{Comparison between proposed eff-IoU and Pixel-F1}
\label{tab:performance_comparison_eff_pixel_f1}
\end{table*}

\section{Effect of Input Noise}
This study evaluates the impact of input Gaussian noise on model performance during inference. 
Gaussian noise with varying intensities, controlled by the noise variance 
[0, 5, 10, 20], is added to the input images as illustrated in 
Fig.~\ref{fig:noise_image}. Since the goal of this experiment is to analyze 
the robustness of the vision-language model (VLM), we exclude the segmentation 
model from this analysis to ensure that only the VLM's performance is evaluated. 
Consequently, evaluation is based solely on the bounding-box IoU predicted by 
the VLM during inference. For this reason, the results reported in this section 
are not directly comparable to the values presented in Section~4.

As shown in Fig.~\ref{fig:noise_image}, our method achieves performance comparable to baseline models, particularly with respect to the eff-IoU metric. The model maintains competitive robustness under low to moderate noise levels. However, the performance advantage gradually diminishes as the noise intensity increases.

\section{Implementation Details}

We implement our method in PyTorch and train it using the Hugging Face Transformers ecosystem with Qwen2.5-VL as the base vision–language model. The model is optimized for image tampering localization, where the target output is either a bounding box in the format $[x, y, w, h]$ indicating the manipulated region or the token $[NO\_EDIT]$ for authentic images.

For data preparation, all images are resized to a fixed spatial resolution of $500 \times 500$ and converted into a unified conversational format suitable for instruction-tuned multimodal models. Each training example contains an input image, a ground-truth manipulation mask, and a textual prompt asking the model to determine whether the image has been edited and, if so, to localize the manipulated region. Ground-truth masks are converted into bounding boxes during preprocessing. Training data is constructed from multiple publicly available image editing and tampering datasets, and we additionally support merged training splits that combine several datasets into a single GRPO-formatted JSON file.

Training is performed using DeepSpeed-based distributed optimization across four NVIDIA H100 GPUs for four epochs. We use a per-device batch size of 8 and sample 8 generations per prompt during GRPO optimization. The model is optimized using AdamW with $\beta_1=0.9$, $\beta_2=0.999$, and $\epsilon=10^{-8}$. The learning rate is set to $5\times10^{-6}$ with a weight decay of 0.1. The implementation exposes configurable parameters including batch size, number of epochs, number of sampled generations, maximum prompt length, and maximum completion length.

The GRPO training objective uses two reward functions: an accuracy reward that evaluates whether the predicted bounding box correctly localizes the manipulated region, and a format reward that encourages the model to produce outputs in the required structured format.

For the segmentation stage, the GRPO-trained VLM is frozen, and its predicted reasoning trace and bounding box are used as prompt inputs. The segmentation module is trained using ground-truth tamper masks. The SAM backbone is kept frozen, and only the reasoning-guided prompt encoder and projection module are optimized.

During inference, the model is prompted to return responses enclosed in structured tags, with the final prediction extracted from the \texttt{<answer>} field. Evaluation reports both localization-oriented IoU and classification-style accuracy metrics based on whether the model correctly predicts edited versus non-edited images.

\section{Evaluation details}
\begin{figure*}[h]
    \centering
    \includegraphics[width=\textwidth]{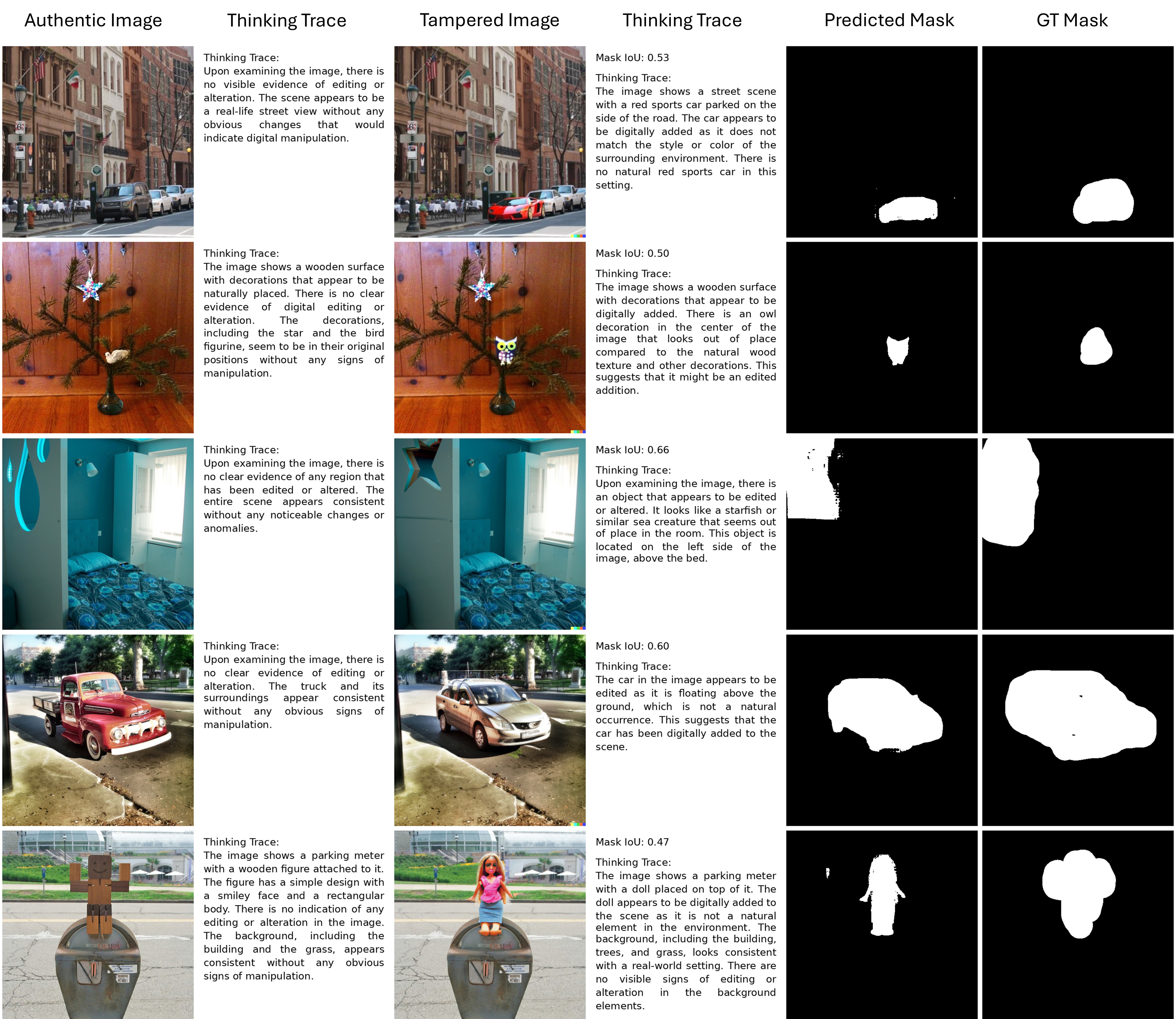}
    \caption{Visualization of model predictions and reasoning traces on the MagicBrush dataset. 
    From left to right: authentic image, reasoning trace for the authentic image, corresponding 
    tampered image, reasoning trace for the tampered image, predicted mask, and ground-truth mask.}
    \label{fig:trace_magicbrush}
\end{figure*}

\begin{figure*}[h]
    \centering
    \includegraphics[width=\textwidth]{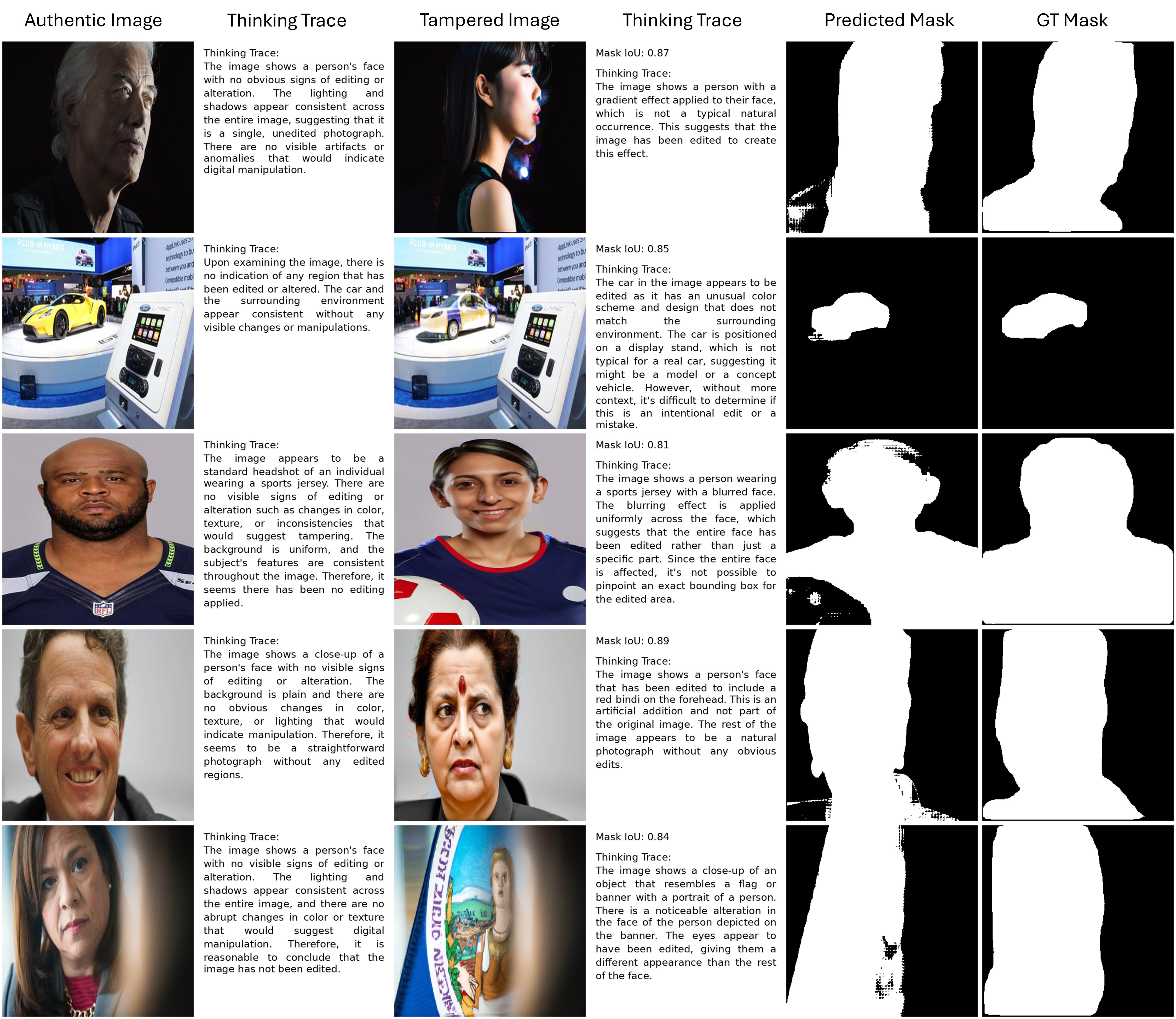}
    \caption{Visualization of model predictions and reasoning traces on the Autosplice dataset. 
    From left to right: authentic image, reasoning trace for the authentic image, corresponding 
    tampered image, reasoning trace for the tampered image, predicted mask, and ground-truth mask.}
    \label{fig:trace_autosplice}
\end{figure*}

\begin{figure*}[h]
    \centering
    \includegraphics[width=\textwidth]{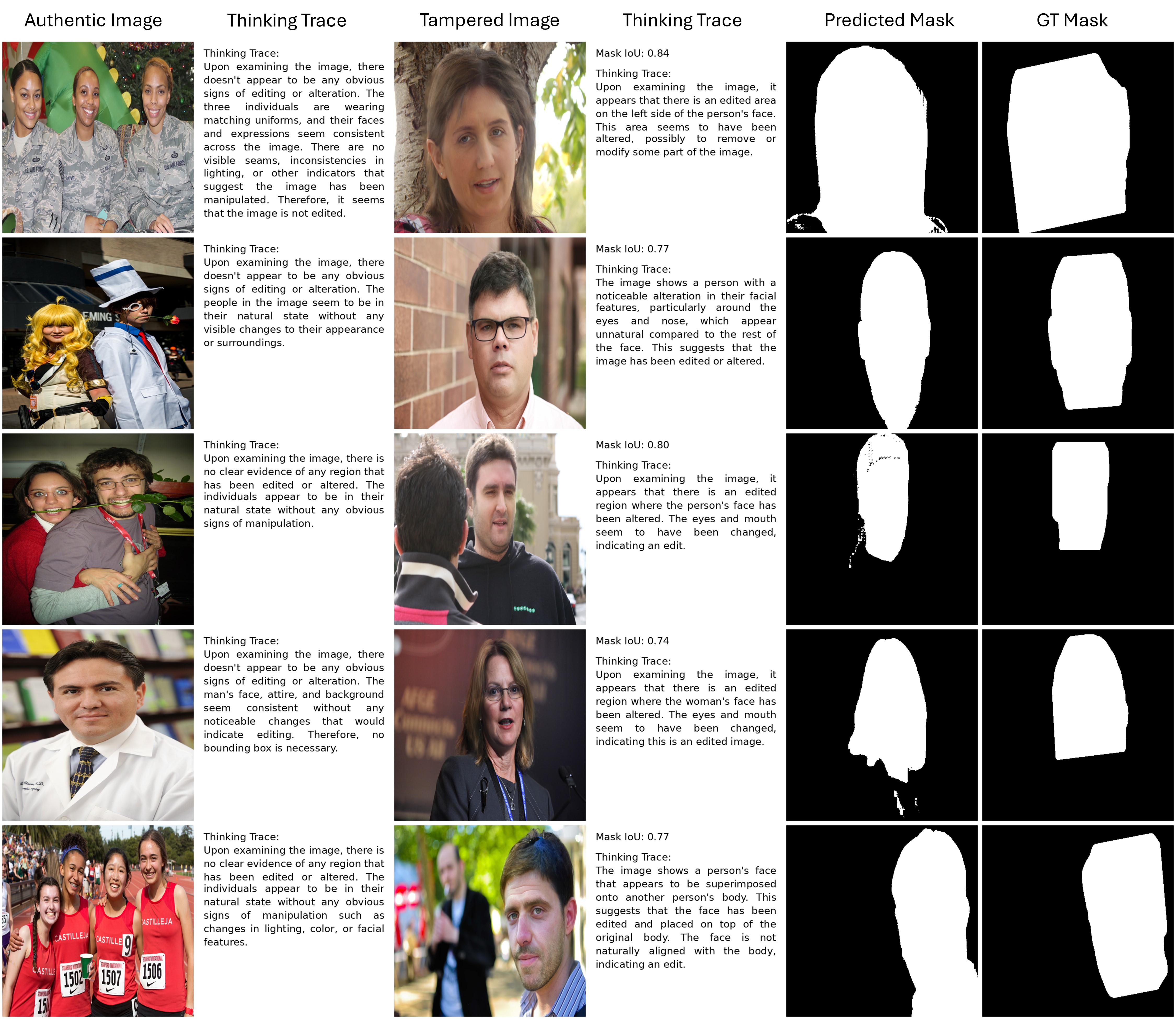}
    \caption{Visualization of model predictions and reasoning traces on the FFHQ-FM dataset. 
    Since FFHQ-FM does not provide one-to-one authentic–tampered image pairs, the authentic 
    and tampered examples shown are selected independently. From left to right: authentic image, 
    reasoning trace for the authentic image, tampered image, reasoning trace for the tampered 
    image, predicted mask, and ground-truth mask.}
    \label{fig:trace_ffhqfm}
\end{figure*}

\begin{figure*}[h]
    \centering
    \includegraphics[width=\textwidth]{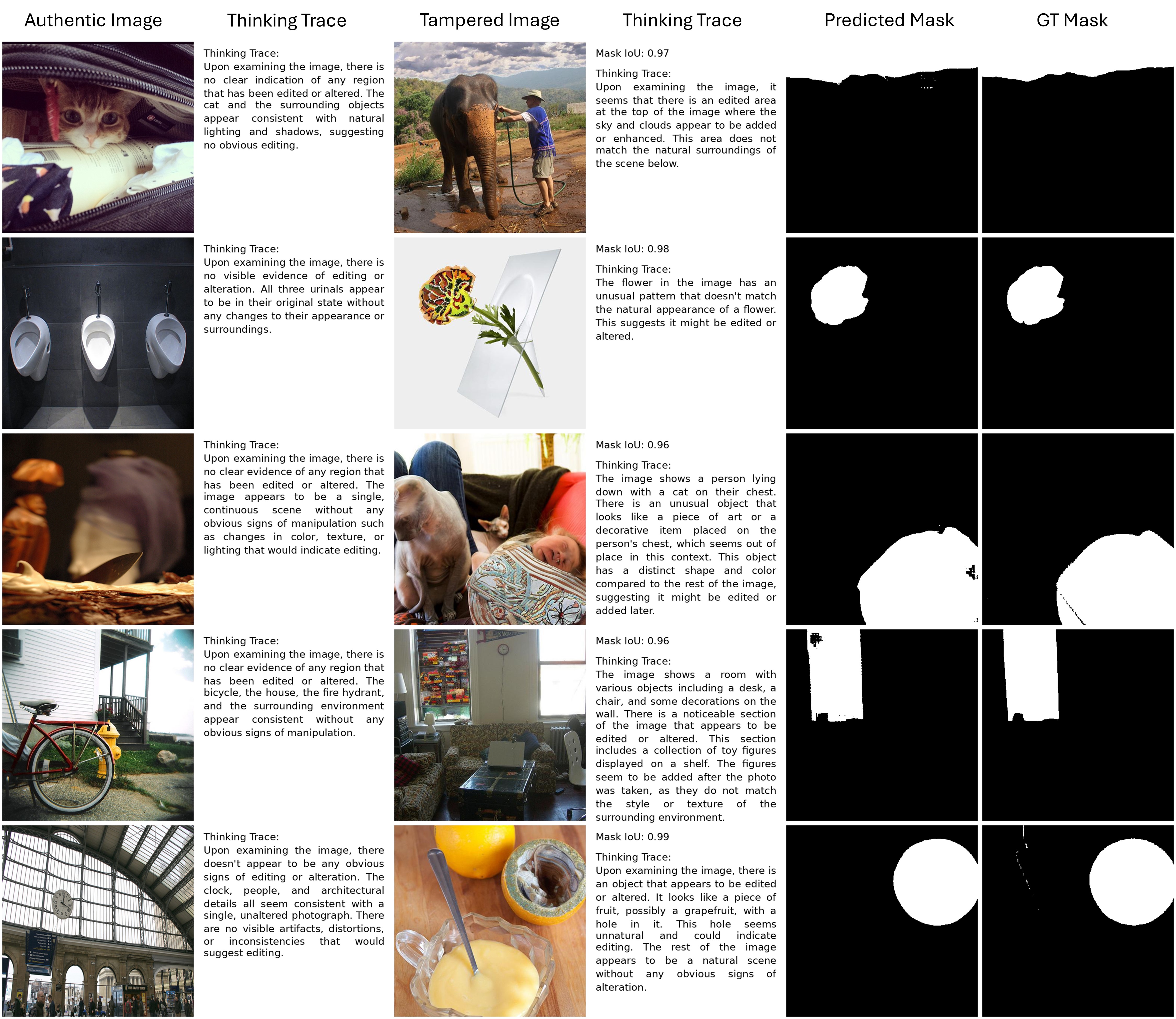}
    \caption{Visualization of model predictions and reasoning traces on the SD\_inpaint dataset. 
    Since SD\_inpaint does not provide one-to-one authentic–tampered image pairs, the authentic 
    and tampered examples shown are selected independently. From left to right: authentic image, 
    reasoning trace for the authentic image, tampered image, reasoning trace for the tampered 
    image, predicted mask, and ground-truth mask.}
    \label{fig:trace_sd_inpaint}
\end{figure*}

\begin{figure*}[h]
    \centering
    \includegraphics[width=\textwidth]{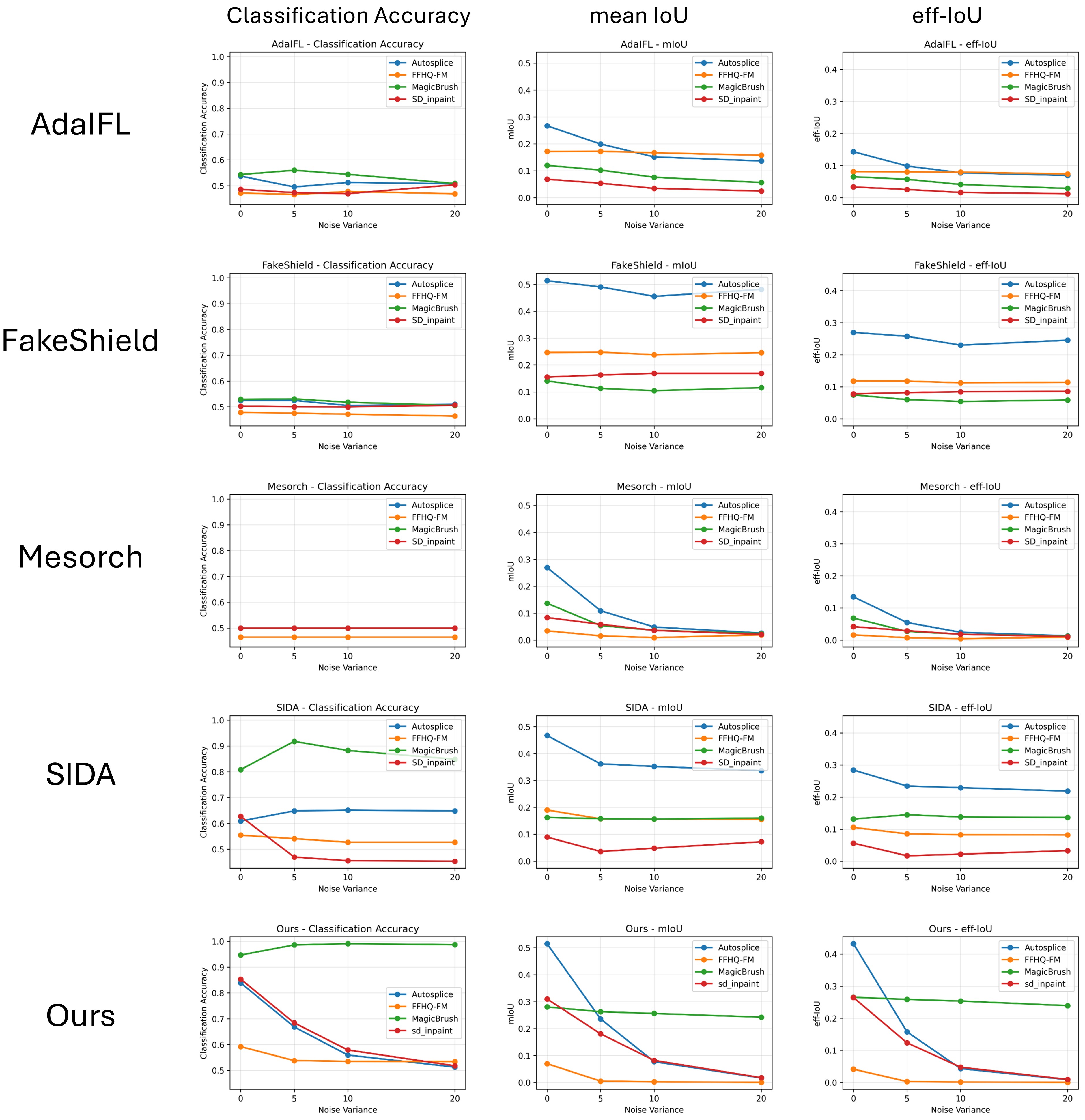}
    \caption{Effect of input Gaussian noise during inference on baseline models. 
    Rows correspond to baseline methods, while columns show the three evaluation 
    metrics: classification accuracy, mIoU, and eff-IoU from left to right. All results are computed 
    across the four datasets used in our experiments.}
    \label{fig:noise_image}
\end{figure*}

In our experiments, all methods are evaluated on the same held-out test splits to ensure a consistent comparison. Almost all baselines share parts of our training data, such as CASIAv2 and Fantastic Reality. In addition, some baselines are trained on the training splits of our evaluation datasets: FakeShield is trained on FFHQ-FM and SD\_inpaint, and SIDA incorporates MagicBrush. Therefore, these cases are valid in-domain comparisons. Also, Tab.~1 in the main paper shows, increase of performance for these in-domain cases are consistent with out of domain cases. Also, we train only using 97K data samples while baselines' training sets range from 106K to 300K+ images. For realistic edits, MagicBrush examples in Fig. \ref{fig:trace_magicbrush} show that the model can detect subtle edits where semantic priors alone are insufficient; Fig.~1 in the main paper also shows in-the-wild examples. 

\end{document}